\documentclass[]{style}

\definecolor{lightblue}{rgb}{0.0, 0.6, 1.0}
\definecolor{darkgreen}{rgb}{0.0, 0.6, 0.0}
\definecolor{lightgreen}{rgb}{0.1, 0.8, 0.1}
\definecolor{lightred}{rgb}{0.7, 0.7, 0.7}
\definecolor{lightgreen}{rgb}{0, 0, 0}
\definecolor{lightlightgray}{rgb}{0.8, 0.8, 0.8}

\usepackage{titlesec}
\titlespacing*{\paragraph}
{0pt}{0em}{0.2em}  

\makeatletter
\renewcommand\paragraph{\@startsection{paragraph}{4}{\z@}%
  {3pt}
  {-0.5em}
  {\normalfont\normalsize\bfseries}} 
\makeatother

\usepackage{amsmath}  
\usepackage{upgreek}
\usepackage{array}    
\usepackage{multirow}
\usepackage{array}
\usepackage{algorithm}
\usepackage{makecell}
\usepackage{graphicx}
\usepackage{algorithm}
\usepackage{algpseudocode}
\usepackage{colortbl}
\usepackage{enumitem}
\usepackage{wrapfig}
\usepackage{mathrsfs}
\usepackage{pifont}
\usepackage{soul}
\usepackage{amssymb}
\usepackage{tcolorbox}
\usepackage{dirtytalk}
\usepackage{xspace}
\usepackage[utf8]{inputenc}

\usepackage{booktabs}

\usepackage{url}
\definecolor{blue}{rgb}{0.21,0.49,0.74}
\definecolor{red}{rgb}{0.8, 0.2, 0.2}
\definecolor{green}{rgb}{0, 0.5, 0}
\definecolor{yellow}{RGB}{218, 160, 109}
\definecolor{gray}{RGB}{155, 155, 155}

\crefname{section}{Sec.}{Secs.}
\Crefname{section}{Section}{Sections}
\Crefname{table}{Table}{Tables}
\crefname{table}{Tab.}{Tabs.}
\crefname{figure}{Fig.}{Figs.}
\Crefname{figure}{Figure}{Figures}
\crefname{appendix}{App.}{Apps.}
\Crefname{appendix}{Appendix}{Appendices}

\usepackage[utf8]{inputenc} 
\usepackage[T1]{fontenc}    
\usepackage{graphicx}
\usepackage{amsmath}
\usepackage{amssymb}
\usepackage{url}            
\usepackage{booktabs}       
\usepackage{amsfonts}       
\usepackage{nicefrac}       
\usepackage{microtype}      
\usepackage{xcolor}         
\usepackage{multirow}
\usepackage{comment}
\usepackage{colortbl}
\usepackage{tocloft}
\usepackage{algorithm}
\usepackage{algorithmicx}
\usepackage{algpseudocode}
\usepackage{wrapfig}
\usepackage{tabularx}
\usepackage{textcomp}
\usepackage{stfloats}
\usepackage{verbatim}
\usepackage{titlesec}
\usepackage{adjustbox}
\usepackage{pifont}
\usepackage{tikz}
\usepackage{color}
\usepackage{subcaption}
\usepackage{soul}

\usepackage{makecell}
\newcolumntype{Y}{>{\centering\arraybackslash}X}
\usepackage{multirow}
\usepackage[table]{xcolor} 
\definecolor{bestcolor}{HTML}{A9D18E} 
\definecolor{sbestcolor}{HTML}{E2EFDA}

\newcolumntype{C}{>{\centering\arraybackslash}m{1.3cm}}
\newcolumntype{Z}{>{\raggedright\arraybackslash\fontsize{6.4pt}{7.2pt}\selectfont}X}

\RequirePackage{xspace}
\makeatletter
\DeclareRobustCommand\onedot{\futurelet\@let@token\@onedot}
\def\@onedot{\ifx\@let@token.\else.\null\fi\xspace}

\makeatother

\definecolor{lightblue}{rgb}{0.66, 0.85, 0.95}
\hypersetup{
    colorlinks=true,
    linkcolor=red,
    citecolor=blue,
}
\definecolor{c2}{HTML}{FBD9BD}
\definecolor{c3}{HTML}{fe793d}
\definecolor{c4}{HTML}{eedeb0}
\definecolor{rouse}{rgb}{0.981,0.961,0.941}
\usepackage[most]{tcolorbox}
\tcbset{colback=blue!5!white, colframe=blue!20!white, boxrule=0pt, arc=2mm, left=1mm, right=1mm, top=1mm, bottom=1mm}

\definecolor{adptorange}{RGB}{248, 205, 172}
\definecolor{cmpblue}{RGB}{189, 215, 238}
\definecolor{cmpblue}{RGB}{189, 215, 238}

\definecolor{our_red}{RGB}{232,157,160}
\definecolor{our_blue}{RGB}{136,206,230}
\definecolor{our_orange}{RGB}{246,200,168}
\definecolor{our_green}{RGB}{178,211,164}

\definecolor{attn_code0}{RGB}{247,215,200}
\definecolor{attn_code1}{RGB}{238,169,139}
\definecolor{mlp_code0}{RGB}{204,201,221}
\definecolor{mlp_code1}{RGB}{102,95,153}

\definecolor{token_blue}{RGB}{84, 120, 140}

\usepackage{pifont}       
\usepackage{bbding}       
\usepackage{fontawesome}
\usepackage{xspace}

\usepackage{float}

\newlength\savewidth

\newcolumntype{x}[1]{>{\centering\arraybackslash}p{#1pt}}
\newcolumntype{y}[1]{>{\raggedright\arraybackslash}p{#1pt}}
\newcolumntype{z}[1]{>{\raggedleft\arraybackslash}p{#1pt}}

\renewcommand{\paragraph}[1]{\vspace{1mm}\noindent\textbf{#1}}

\usepackage{colortbl}
\usepackage{xcolor}
\usepackage{wrapfig}

\renewcommand{\paragraph}[1]{\vspace{1.25mm}\noindent\textbf{#1}}

\usepackage{listings}

\definecolor{codeblue}{rgb}{0.21, 0.49, 0.74}
\definecolor{codekw}{rgb}{0.35, 0.35, 0.75}
\lstdefinestyle{Pytorch}{
    language = Python,
    backgroundcolor = \color{white},
    basicstyle = \fontsize{9pt}{8pt}\selectfont\ttfamily\bfseries,
    columns = fullflexible,
    aboveskip=1pt,
    belowskip=1pt,
    breaklines = true,
    captionpos = b,
    commentstyle = \color{codeblue},
    keywordstyle = \color{codekw},
}

\definecolor{green}{HTML}{009000}
\definecolor{red}{HTML}{ea4335}

\title{SmartDirector: Keyframe-Conditioned Cinematic Video Generation with Narrative Pacing Control}
\author[1]{Zhida Zhang}
\author[2]{Jie Ma}
\author[3]{Zhan Peng}
\author[2]{Haoxue Wu}
\author[2]{Yang Han}
\author[2]{Jun Liang}
\author[1]{Jie Cao}
\author[2]{Jing Li}

\affiliation[1]{NLPR, CISIA}
\affiliation[2]{Youku Moku-Lab}
\affiliation[3]{HUST}

\abstract{

The narrative quality of a video fundamentally determines its perceptual value. Although existing video generation methods can produce visually appealing content, they predominantly rely on sparse conditioning signals such as text prompts or first/last frames, which limits precise control over narrative structure and temporal pacing. 
In this paper, we propose SmartDirector, a framework that enhances the narrative capacity of video generation models through multiple keyframes. SmartDirector supports flexible generation scenarios including single-shot generation, multi-shot narrative synthesis, and video extension. The framework operates in two stages: Director-Gen generates a low-resolution video conditioned on the provided keyframes, and Director-SR refines the output by exploiting high-resolution keyframes as semantic anchors to recover fine-grained details. To enable robust multi-keyframe training, we construct a data pipeline that curates single-shot and multi-shot sequences from movies. Extensive experiments demonstrate that SmartDirector substantially outperforms existing state-of-the-art approaches. We will release the code to facilitate further research.

}

\project{\url{https://orange-3dv-team.github.io/SmartDirector/}}
\date{\today}

\begin{document}
\thispagestyle{firstheader}
\maketitle
\pagestyle{plain}

\section{Introduction}
\label{sec:intro}

\begin{figure}[t]
    \centering
    \includegraphics[width=0.99\linewidth]{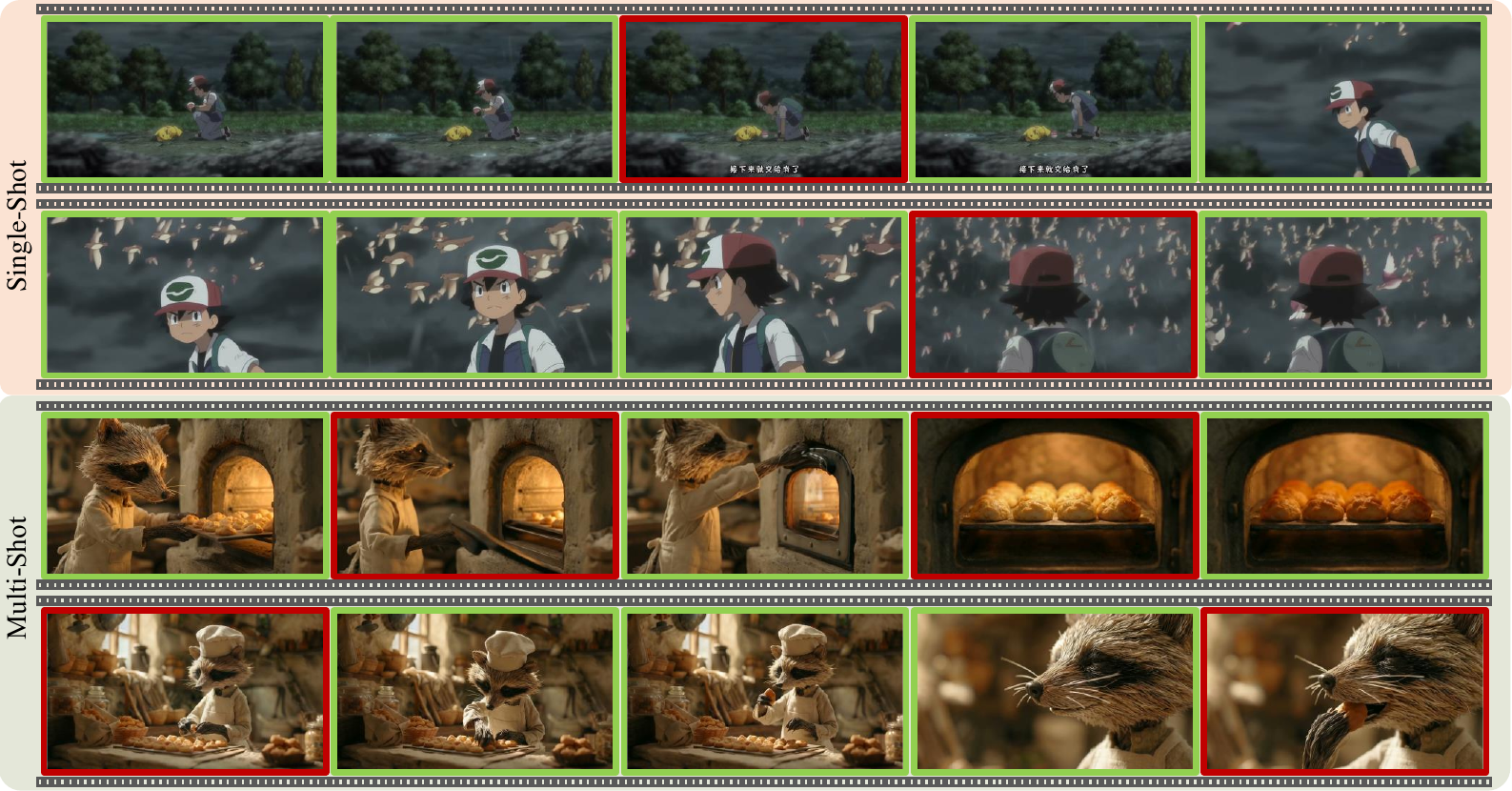}
    \caption{\textbf{Examples generated by SmartDirector.} SmartDirector enables high-fidelity video generation guided by arbitrary keyframes.}
    \label{fig:teaser}
    \vspace{-16pt}
\end{figure}

Recent advancements in video generation have propelled a paradigm shift from synthesizing short, single-shot clips~\cite{wan2025wanopenadvancedlargescale,kong2024hunyuanvideo,HaCohen2024LTXVideo} to creating long, multi-shot narratives~\cite{wang2025multishotmaster,klingteam2025klingomnitechnicalreport,meng2025holocine,sora,veo,xiao2025captain}. 
Although existing methods have demonstrated remarkable capabilities in generating visually stunning and high-fidelity videos, they predominantly rely on sparse conditioning signals, such as text prompts or first/last frames. Consequently, these approaches struggle to achieve precise control over fine-grained spatial-temporal content and narrative structure, significantly restricting their practical utility in real-world applications. 
In professional filmmaking, directors use storyboards~\cite{wiki:storyboard} to guide the production process and exercise fine-grained control over visual content. Storyboards serve as visual anchors that maintain coherence across multiple shots and regulate the temporal pacing (i.e., the rhythm and timing of visual content) within each individual shot. In this work, we identify keyframes as the direct counterpart of storyboards in video generation. Building on this perspective, we focus on the task of multi-keyframe-conditioned video generation.

A naive approach is to treat each pair of adjacent keyframes as the start and end frames of a short clip, generate the clips autoregressively, and concatenate the results. However, this strategy neglects global context during synthesis, resulting in abrupt temporal discontinuities at keyframe boundaries and a loss of narrative consistency across the entire video.
Recent work~\cite{liu2025pusa,liu2025dreamontage} proposes an alternative that inserts keyframes directly into noisy latents at their corresponding temporal positions before denoising with a video diffusion model. Yet this method is fundamentally limited by the causal structure of the temporal VAE~\cite{wan2025wanopenadvancedlargescale,kong2024hunyuanvideo,yang2024cogvideox}. In a standard 3D VAE, the first frame is encoded independently, while subsequent frames are encoded in groups (e.g., every four frames) with causal dependence on preceding frames. Direct latent replacement at arbitrary positions violates this causal dependency, producing temporal discontinuities and visual artifacts near the keyframes.

\begin{wrapfigure}{r}{0.48\textwidth}
    \centering
    \vspace{-8pt}
    \includegraphics[width=\linewidth]{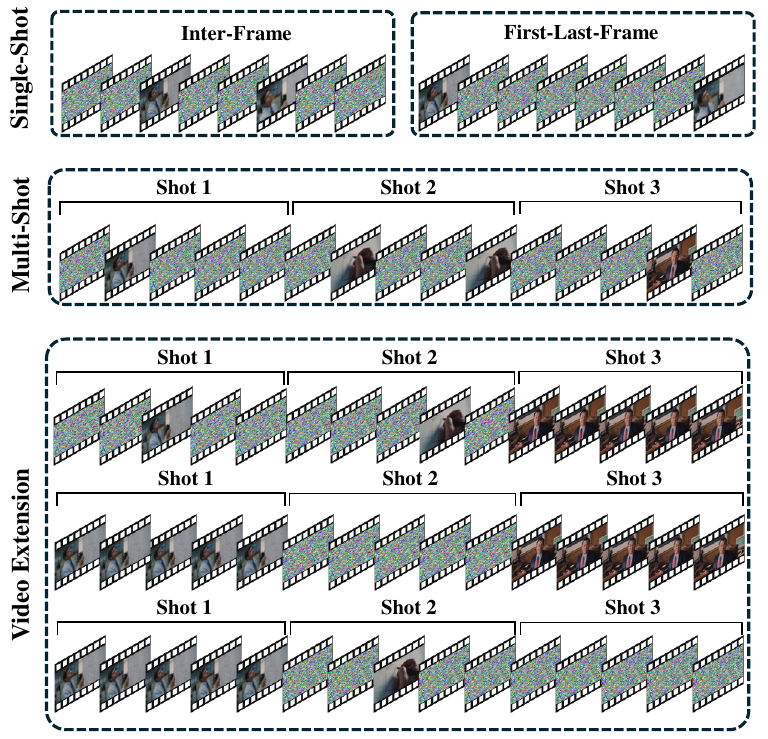}
    \vspace{-12pt}
    \caption{\textbf{SmartDirector is a flexible framework} that accommodates diverse input conditions and supports a wide range of generation tasks, including first-frame-to-video, last-frame-to-video, multi-shot synthesis, and video extension.}
    \vspace{-5pt}
    \label{fig:fig2}
\end{wrapfigure}

In this paper, we introduce SmartDirector, a flexible framework for video generation guided by arbitrary keyframes that seamlessly supports both single-shot and multi-shot synthesis. Beyond keyframe-conditioned generation, SmartDirector also supports video-conditioned generation for video extension, as illustrated in Fig.~\ref{fig:fig2}. To fully exploit the conditioning provided by multiple keyframes, the framework consists of two stages: a keyframe-conditioned generation stage and a keyframe-conditioned super-resolution stage, referred to as Director-Gen and Director-SR.
In the Director-Gen stage, we propose a Multi-Chunk VAE strategy to address the causal limitation of the temporal VAE. During training, the video is partitioned into multiple chunks at the keyframe positions, with each keyframe serving as the first frame of its respective chunk and encoded independently by the VAE. The resulting multi-chunk latents are then processed by a Diffusion Transformer (DiT)~\cite{peebles2023scalable}. To maintain global consistency, we apply full spatio-temporal attention within the DiT, enabling each chunk to attend to the global context across all chunks.
Videos produced by the Director-Gen stage are typically low-resolution (e.g., 480p), which is lower than the resolution of the provided keyframes. To leverage the fine-grained details in the high-resolution keyframes, we design a keyframe-conditioned super-resolution module in the Director-SR stage that upsamples the generated video to high definition (e.g., 1080p), explicitly conditioned on the high-resolution keyframes.

Training our framework requires carefully curated data. We construct a data processing pipeline for curating long video sequences, as illustrated in Fig.~\ref{fig:datapipe}. For the Director-Gen stage, we collect copyright-free movies and segment them into single shots. We then compute visual similarities between these shots to aggregate them into coherent multi-shot video sequences, which are further annotated with structured descriptions using Vision-Language Models (VLMs)~\cite{Qwen3-VL,team2023gemini}. The resulting dataset contains both single-shot and multi-shot sequences, enabling robust training for both generation settings. For the super-resolution task, we use the open-source UltraVideo dataset~\cite{xue2025ultravideo}.
Our main contributions are summarized as follows:
\begin{itemize}
    \item We propose SmartDirector, a unified framework that enables flexible keyframe-conditioned video generation, covering single-shot, multi-shot, and video extension.
    \item We identify the fundamental limitation imposed by the causal structure of the temporal VAE on keyframe insertion and propose a Multi-Chunk VAE strategy. This design circumvents the causal constraints, allowing keyframes to be placed at arbitrary temporal positions while ensuring smooth and continuous generation.
    \item We design a keyframe-conditioned super-resolution module that exploits high-resolution keyframes as semantic anchors to recover fine-grained details.
\end{itemize}

\section{Related Work}
\label{sec2:related}

\subsection{Video Generation}
Video generation has evolved rapidly from synthesizing short single-shot clips~\cite{wan2025wanopenadvancedlargescale,kong2024hunyuanvideo,HaCohen2024LTXVideo} to producing long multi-shot narratives~\cite{wang2025multishotmaster,klingteam2025klingomnitechnicalreport,meng2025holocine,sora,veo,xiao2025captain}.
However, these methods rely on sparse conditioning signals such as text prompts or the first/last frame, which limits their ability to control fine-grained spatial-temporal content and narrative structure.
Recently, several approaches have attempted to incorporate multiple keyframes into the generation process to enable more precise control.
For single-shot video generation, Pusa~\cite{liu2025pusa} injects noise of different timesteps into distinct frames, while DreaMontage~\cite{liu2025dreamontage} directly inserts keyframes into noisy latents at corresponding positions.
For multi-shot video generation, CaptainCinema~\cite{xiao2025captain} generates video by conditioning on the first frames of each shot.
However, these methods are limited to specific scenarios and lack the flexibility to support arbitrary keyframe placement for precise temporal and spatial control.
In this work, we propose a unified framework that enables flexible keyframe control for both single-shot and multi-shot video generation.
Additionally, our method supports video extension by using video frames as input to extend the content temporally.

\subsection{Video Super Resolution}
Video super-resolution has been studied extensively over the past decades.
Early approaches were predominantly based on GANs~\cite{chu2018temporally,chan2022investigating}, while recent methods have shifted toward diffusion models~\cite{wang2025seedvr,chen2025dove,yu2026sparkvsr}.
SeedVR~\cite{wang2025seedvr} introduces a shifted window attention mechanism to enable effective restoration on long video sequences.
DoVE~\cite{chen2025dove} proposes an efficient one-step diffusion model for real-world video super-resolution.
However, existing VSR methods primarily focus on pixel-level enhancement, often treating each frame as a restoration target rather than a semantic object to be reconstructed.
Consequently, they struggle to address common artifacts in low-resolution videos generated in the first stage, such as distorted small faces and incorrect text.
A concurrent work, SparkVSR~\cite{yu2026sparkvsr}, also explores keyframe-conditioned video super-resolution.
In contrast, our Director-SR is designed as the refinement stage of a unified keyframe-conditioned generation framework, using multiple high-resolution keyframes as semantic anchors to reconstruct fine-grained details and correct generative artifacts throughout the sequence.

\section{Method}

\begin{figure}[t]
    \centering
    \includegraphics[width=\linewidth]{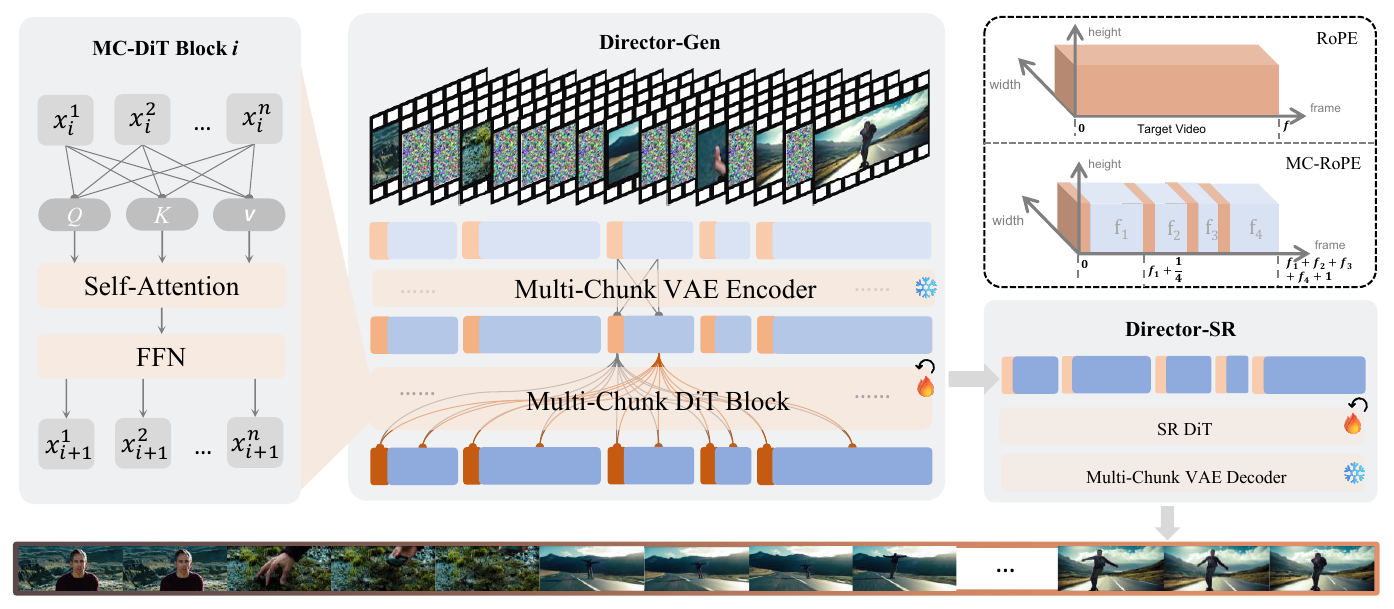}
    \caption{\textbf{Overview of the proposed SmartDirector framework.} SmartDirector is a two-stage framework for keyframe-conditioned video generation. In the first stage, Director-Gen synthesizes a low-resolution video from the provided keyframes via a Multi-Chunk VAE that encodes chunks independently at keyframe boundaries and a Multi-Chunk DiT with full spatio-temporal attention and MC-RoPE for coherent cross-chunk modeling. In the second stage, Director-SR refines the low-resolution output into a high-resolution video under the guidance of high-resolution keyframes.}
    \vspace{-8pt}
    \label{fig:framework}
\end{figure}

SmartDirector is a two-stage framework that comprises a keyframe-conditioned generation stage (Director-Gen) and a keyframe-conditioned super-resolution stage (Director-SR), as illustrated in Fig.~\ref{fig:framework}. This section first provides a brief review of the Flow Matching framework, then details the proposed method, and finally describes the data curation pipeline.

\subsection{Flow Matching}

Let $z_0 \sim Z_0$ denote a data sample and $z_1 \sim Z_1$ denote a noise sample. Recent image generation models (e.g., \cite{esser2024scaling, flux2024}) and video generation models (e.g., \cite{wan2025wanopenadvancedlargescale, sora, kong2024hunyuanvideo, chen2025goku}) adopt the Rectified Flow \cite{liu2022flow} framework, which defines the interpolated latent $z_t$ as
\begin{equation}\label{add_noise}
z_t = (1 - t) z_0 + t z_1,
\end{equation}
for $t \in [0, 1]$. The model is trained to regress the velocity field $\boldsymbol{v}_\theta(z_t, t)$ by minimizing the Flow Matching objective~\cite{lipman2022flow}:
\begin{equation}
\mathcal{L}(\theta) = \mathbb{E}_{t, z_0, z_1} \left[ \| \boldsymbol{v} - \boldsymbol{v}_\theta(z_t, t) \|^2 \right],
\end{equation}
where the target velocity field is $\boldsymbol{v} = z_1 - z_0$.

\subsection{Director-Gen}

Training Director-Gen requires a set of videos, each paired with a structured caption $c$ and a set of keyframes $\{I_k\}$ ($k$ denotes the keyframe index). To address the causal limitation of the 3D VAE, we propose a Multi-Chunk VAE strategy.

We first split the video $V$ into $n$ video chunks $\{V_j\}$ ($j$ denotes the chunk index) at the keyframe positions, ensuring that each keyframe serves as the first frame of its respective chunk. For simplicity, we assume the first frame is always provided as a keyframe (i.e., $I_0$), so the number of chunks equals the number of keyframes. During training, noise is injected exclusively into non-keyframe positions, following Eq.~\ref{add_noise}. We then encode these chunks with a 3D causal VAE $\mathcal{E}$ for spatiotemporal compression, yielding latents $z_j = \mathcal{E}(V_j)$. Through this design, each keyframe is encoded independently by the VAE.
These chunk latents are patchified into visual tokens $\{x_j \in \mathbb{R}^{f_j \times d \times h \times w}\}$, where $f_j$, $d$, $h$, and $w$ denote the latent frame count, channel number, height, and width, respectively. We concatenate all chunk tokens along the temporal dimension to form a unified latent sequence:
\begin{equation}
    x = \mathrm{Concat}(\{x_j\}), \quad x \in \mathbb{R}^{(\sum f_j) \times d \times h \times w}.
\end{equation}
The sequence $x$ is then processed by the DiT. To maintain global consistency, we apply full spatio-temporal attention across all chunks.

The DiT employs 3D Rotary Positional Embeddings (RoPE) to encode spatiotemporal coordinates, where temporal indices typically increment sequentially as non-negative integers. However, applying a single continuous temporal frame index across the unified multi-chunk latent $x$, or resetting the temporal frame index for each chunk latent $x_j$, introduces temporal discontinuities at keyframe boundaries. To address this, we propose \textbf{Multi-Chunk RoPE (MC-RoPE)}, which assigns fractional temporal indices to keyframe positions, thereby preserving temporal smoothness across chunk boundaries. 
Specifically, the temporal index $u_i$ for latent $x$ is computed as:
\begin{equation}
u_i = 
\begin{cases} 
    u_{i-1} + 1, & \text{if } i \neq \hat{k}, \\ 
    u_{i-1} + 0.25, & \text{if } i = \hat{k}, 
\end{cases}
\end{equation}
where $i$ denotes the latent temporal index, $\hat{k}$ denotes the keyframe index in latent, and $u_0=0$.

\subsection{Director-SR}
Videos produced by the Director-Gen stage are low-resolution (e.g., 480p) due to the computational cost of diffusion models. Consequently, they struggle to preserve fine details such as facial features and text, limiting their practical applicability. Existing video super-resolution (VSR) methods focus on pixel-level restoration and lack precise frame-level control, making them insufficient for correcting generative artifacts introduced in the Director-Gen stage.

To exploit the high-resolution keyframes as semantic anchors, we design a keyframe-conditioned super-resolution module in the Director-SR stage. During training, each sample consists of paired low-resolution (LR) and high-resolution (HR) videos, denoted as $V^{LR}$ and $V^{HR}$. A subset of HR frames is sampled from $V^{HR}$ to serve as keyframes $\{I_k\}$. Following existing methods~\cite{zhuang2025flashvsr,yu2026sparkvsr}, $V^{LR}$ is synthesized by applying degradation operations to $V^{HR}$.
As in the Director-Gen stage, we adopt the Multi-Chunk VAE strategy to circumvent the causal constraints of the VAE, obtaining the HR latents $z^{HR}$ and LR latents $z^{LR}$. 
The LR latent $z^{LR}$ is spatially upsampled to match the spatial dimensions of $z^{HR}$. At the keyframe indices, the LR latents are replaced with the corresponding HR latents to enforce keyframe conditioning. We then use flow matching to predict the velocity field mapping $z^{LR}$ to $z^{HR}$. Following Eq.~\eqref{add_noise}, the interpolation is formulated as:
\begin{equation}
    z_t = (1-t)z^{HR} + tz^{LR}.
\end{equation}
Note that our Director-SR stage is designed to refine the results of the Director-Gen stage, it can also operate independently to perform super-resolution on arbitrary low-resolution videos.

\begin{figure}[t]
    \centering
    \includegraphics[width=\linewidth]{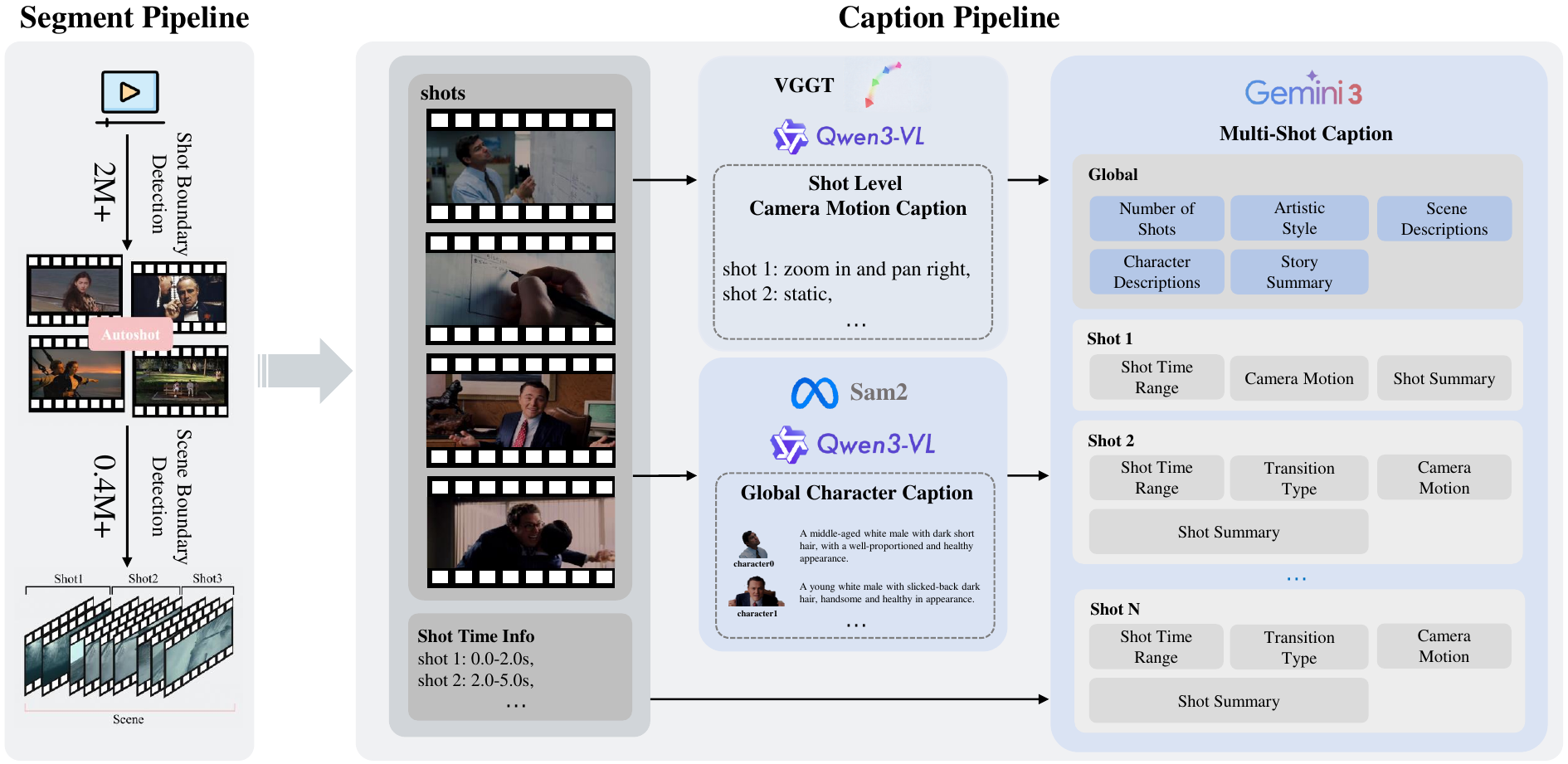}
    \caption{\textbf{Overview of the data pipeline.} Starting from large-scale cinematic videos, we perform shot segmentation with AutoShot and VLM-based multi-shot aggregation, and then generate structured captions that cover camera motion, per-character appearance, and a holistic description of the whole multi-shot sequence together with descriptions of each individual shot.}
    \label{fig:datapipe}
    \vspace{-8pt}
\end{figure}

\subsection{Data Pipeline}
Training SmartDirector requires a large corpus of videos paired with structured captions that describe both the overall narrative and the content of each individual shot. To this end, we build a scalable data pipeline that proceeds in three steps, as illustrated in Fig.~\ref{fig:datapipe}.

\textbf{Video collection and shot segmentation.} We first collect a large-scale set of cinematic videos from publicly available sources. Each raw video is then partitioned into single-shot clips using AutoShot~\cite{zhuautoshot}. To construct multi-shot samples that preserve narrative continuity, we further employ a vision-language model to aggregate consecutive single-shot clips that share the same scene and storyline, yielding multi-shot videos with coherent semantics.

\textbf{Structured video captioning.} We then annotate each video with a systematic, multi-aspect caption. To describe camera behavior, we combine VGGT~\cite{wang2025vggt} for geometric camera trajectory estimation with Qwen3-VL~\cite{Qwen3-VL} for visual interpretation (e.g., pan, zoom, dolly). To characterize on-screen subjects, we track each character with SAM2~\cite{ravi2024sam2} across the entire video and generate an appearance description for every tracked identity via Qwen3-VL, ensuring consistent character grounding across shots.

\textbf{Hierarchical caption aggregation.} Finally, we feed the shot-level visual content, camera descriptions, and character descriptions into Gemini-3-Pro~\cite{team2023gemini} to produce a structured caption in a unified format. The caption contains a holistic description that summarizes the overall narrative of the multi-shot video, as well as per-shot descriptions that specify the visual content, camera motion, and active characters of each individual shot.

\section{Experiments}
\label{sec:experiments}

\begin{table}[t]
\centering
\caption{\textbf{Comparison of FVD and LLM-based evaluation between Dreamina and SmartDirector} on single-shot and multi-shot scenarios. Best values per metric are highlighted in \textbf{bold}.}
\label{tab:combined_eval}
\resizebox{\textwidth}{!}{%
\setlength{\tabcolsep}{6pt}
\renewcommand{\arraystretch}{1.15}
\begin{tabular}{llcccccccc}
\toprule
\multirow{2}{*}{\textbf{Scenario}} & \multirow{2}{*}{\textbf{Method}} & \multirow{2}{*}{\textbf{FVD} $\downarrow$} & \multicolumn{5}{c}{\textbf{LLM-based Evaluation} $\uparrow$} & \multirow{2}{*}{\textbf{Avg.} $\uparrow$} \\
\cmidrule(lr){4-8}
 & & & \textbf{Inst.} & \textbf{Narr.} & \textbf{Phys.} & \textbf{Qual.} & \textbf{Aest.} & \\
\midrule
\multirow{2}{*}{Single-Shot}
 & Dreamina & 226.85 $\pm$ 3.44 & 89.45 & 78.46 & 85.73 & 79.81 & 85.90 & 83.87 \\
 & SmartDirector & \textbf{41.12 $\pm$ 1.01} & \textbf{92.97} & \textbf{91.02} & \textbf{91.24} & \textbf{90.44} & \textbf{90.85} & \textbf{91.30} \\
\midrule
\multirow{2}{*}{Multi-Shot}
 & Dreamina & 251.83 $\pm$ 5.87 & 64.60 & 57.23 & 60.52 & 46.46 & 67.78 & 59.32 \\
 & SmartDirector & \textbf{65.65 $\pm$ 2.46} & \textbf{89.70} & \textbf{86.08} & \textbf{88.75} & \textbf{88.18} & \textbf{89.69} & \textbf{88.48} \\
\bottomrule
\end{tabular}
}
\end{table}
\vspace{-8pt}

\subsection{Experimental Setup}
\textbf{Implementation Details.} 
For the Director-Gen stage, we adopt a 32B internal diffusion model\footnote{The model shares a similar architecture with Wan-2.1-T2V; we plan to release a 14B variant for broader accessibility.} as the base model. For the Director-SR stage, we employ Wan-2.2-5B as the backbone. Both stages share the same Wan-2.2-VAE for latent encoding. We fine-tune all parameters of the DiT in both stages. The Director-Gen model is trained on 40 NVIDIA GPUs for 20,000 steps, while the Director-SR model is trained on 8 NVIDIA GPUs for 2,000 steps. Both stages use a learning rate of $2 \times 10^{-5}$.

\textbf{Benchmark.}
To facilitate a comprehensive quantitative evaluation, we construct a diverse benchmark from movies, TV series, and animations. The benchmark comprises 250 single-shot videos and 250 multi-shot videos, with durations ranging from 3 to 15 seconds. All videos are rendered at 24 FPS with a native resolution of at least 1080p. 
For each video, we randomly sample a set of keyframes as conditioning signals. 
To ensure compatibility with the causal structure of the temporal VAE, the number of frames in each chunk must satisfy $4n + 1$, where $n$ is a non-negative integer.
For the Director-SR stage, we additionally evaluate on existing video super-resolution benchmarks to measure the super-resolution performance independently from the generation stage. As prior work on keyframe-conditioned video generation is scarce, we compare SmartDirector against Dreamina Multiframes~\cite{jimeng2024,liu2025dreamontage}, the most representative closed-source system that supports multi-keyframe conditioning.

\subsection{SmartDirector Results}
\textbf{Metrics.} We evaluate the generated videos from three complementary perspectives. As the basic objective metric, we report FVD to measure the distributional fidelity between generated and real videos. Since FVD primarily reflects low-level statistics, we further employ a vision-language model for high-level semantic assessment, where Gemini-3-Pro~\cite{team2023gemini} scores each video along 5 semantic dimensions (Instruction-Following, Narrative Coherence, Physical Consistency, Video Quality, and Video Aesthetic Quality; full definitions are provided in Appendix~\ref{sec:appendix_llm_eval}). Finally, we conduct a user study following the pairwise Good/Same/Bad (GSB) protocol on four perceptual aspects.

\textbf{Objective Quality.} As shown in Table~\ref{tab:combined_eval}, SmartDirector achieves substantially lower FVD than the baseline across both scenarios. In the Single-Shot setting, our method reduces FVD from 226.85 to 41.12, reflecting a closer distributional match to real videos in terms of per-frame visual quality and temporal dynamics. The gain is even more pronounced in the Multi-Shot setting (251.83 vs.~65.65), where scene transitions and camera cuts introduce additional temporal complexity. 

\textbf{Semantic Fidelity.} To assess high-level semantics beyond pixel-level metrics, we employ a large multimodal model to score generated videos across 5 dimensions. In the single-shot scenario, SmartDirector improves the average score from 83.87 to 91.30, with the largest gains in Narrative Coherence (+12.56). In the multi-shot scenario, the margin widens further: the average score increases from 59.32 to 88.48. These results demonstrate the superior narrative capabilities of SmartDirector.

\begin{wrapfigure}{r}{0.5\textwidth}
    \centering
    \vspace{-10pt}
    \includegraphics[width=\linewidth]{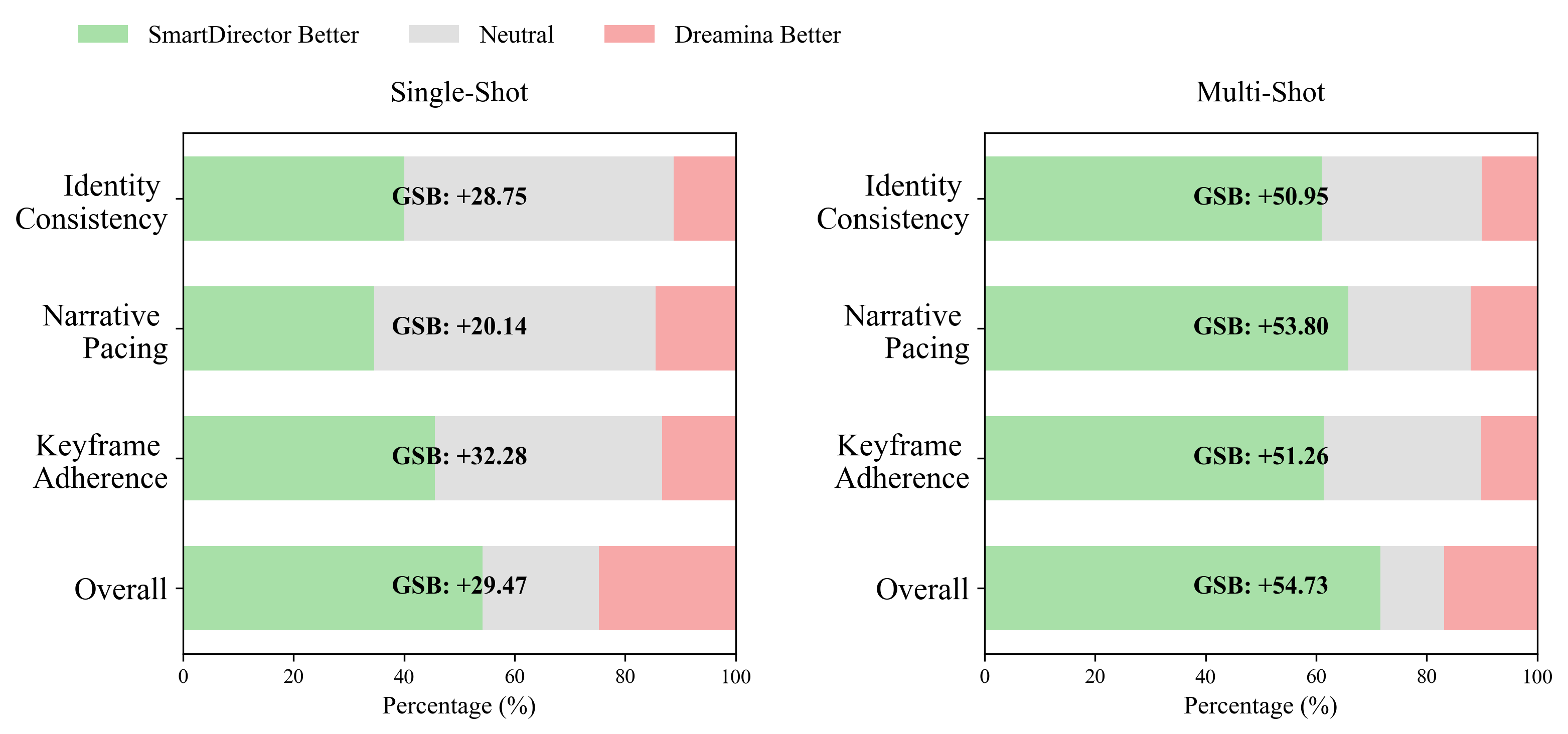}
    \vspace{-12pt}
    \caption{\textbf{Human evaluation (GSB) comparison between SmartDirector and the Dreamina.}}
    \vspace{-5pt}
    \label{fig:human_eval}
\end{wrapfigure}
\textbf{Human evaluation.} To assess perceptual quality, we conduct a user study following the Good/Same/Bad (GSB) protocol. Thirty participants evaluate 500 video pairs generated by SmartDirector and Dreamina across four dimensions: Identity Consistency, Narrative Pacing,  Keyframe Adherence, and Overall Quality. The GSB score is defined as
\begin{equation}
\text{GSB} = \frac{\text{Wins} - \text{Losses}}{\text{Wins} + \text{Losses} + \text{Ties}}.
\end{equation}

As shown in Fig.~\ref{fig:human_eval}, SmartDirector consistently outperforms the baseline across all scenarios. In single-shot settings, our method achieves clear advantages in Narrative Pacing, indicating that keyframe-conditioned generation effectively preserves temporal dynamics. In multi-shot settings, the margin increases further, with a 54.73\% win rate in Overall Quality, demonstrating that SmartDirector mitigates identity drift and narrative fragmentation across shot boundaries. Detailed per-dimension statistics are provided in Appendix~\ref{sec:appendix_human_eval}.

\begin{figure}[htbp]
    \centering
    \vspace{-1em}
    \includegraphics[width=\textwidth]{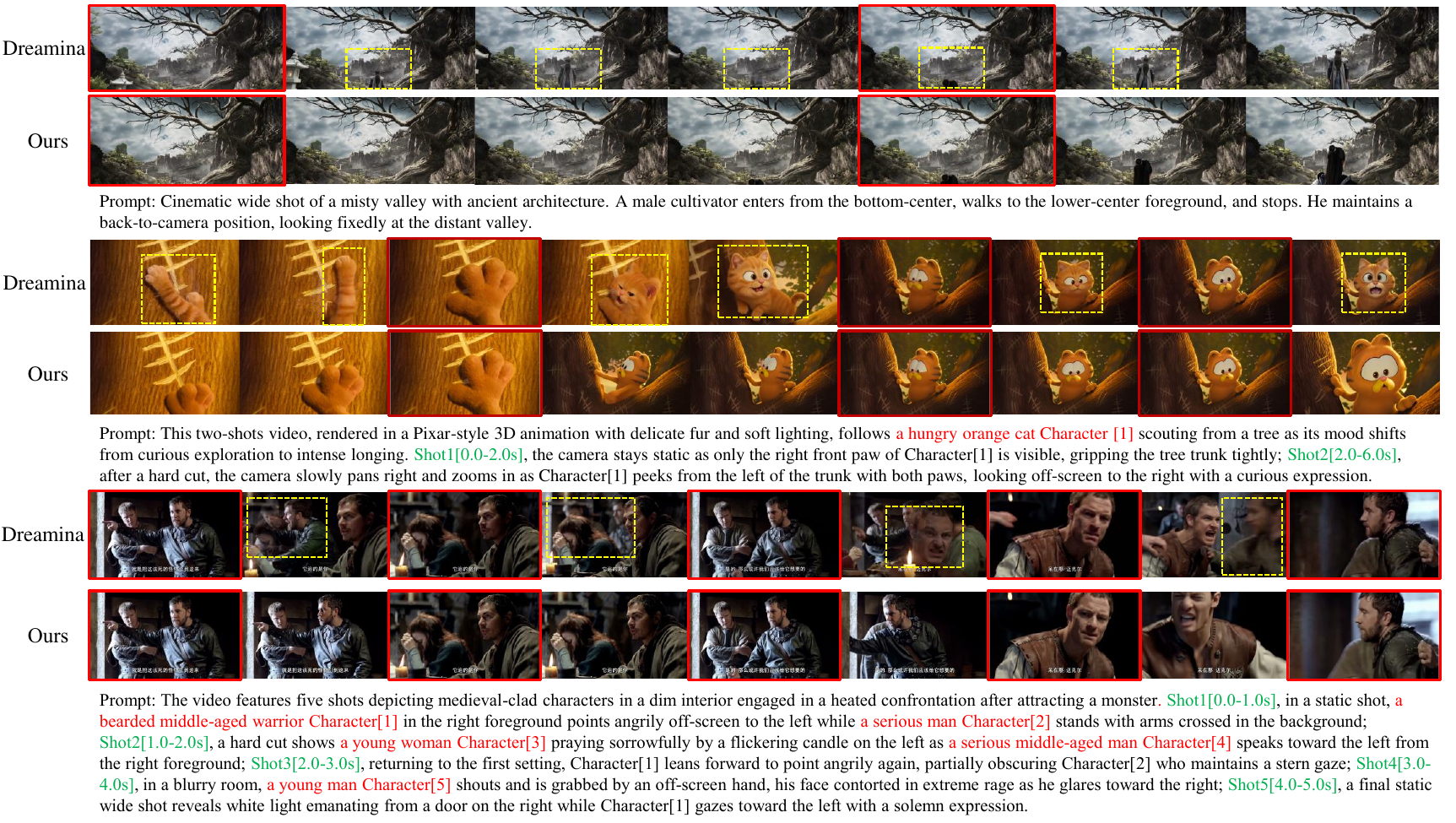}
    \caption{\textbf{Qualitative comparison with Dreamina.} Comparison with state-of-the-art methods. Red box denotes the keyframe, yellow dashed box denotes the generated artifacts.}
    \label{fig:compare}
\end{figure}

\textbf{Qualitative Results. }
We compare SmartDirector with Dreamina~\cite{jimeng2024} on representative examples in Fig.~\ref{fig:compare}. As shown in the figure, Dreamina frequently produces artifacts in the intermediate frames between keyframes, as indicated by the yellow dashed boxes. In the first case, the character exhibits implausible motion trajectories that violate the underlying physical dynamics, while in the second case we observe noticeable identity drift, where the cat's appearance gradually deviates from the appearance specified by the keyframes. Such degradations become even more pronounced in the multi-shot setting (third case), where Dreamina produces flickering characters between shots. 
These failures reflect the model's inability to maintain visual continuity when keyframe intervals exceed its effective temporal receptive field. 
In contrast, SmartDirector generates coherent and contextually consistent intermediate frames across both single-shot and multi-shot settings. The resulting sequences preserve entity identity, spatial layout, and motion dynamics throughout the keyframe intervals without observable artifacts. This improvement stems from our explicit frame-level keyframe conditioning, which aligns each keyframe with the first frame of its corresponding chunk and thereby provides strong temporal anchors for coherent generation.

\subsection{Director-SR Results}
\textbf{Quantitative Results. }We note that although our Director-SR is designed to refine the Director-Gen results, it can also serve as a standalone keyframe-conditioned VSR method.
To isolate the contribution of Director-SR from Director-Gen, we evaluate it independently against SparkVSR.
As shown in Table~\ref{tab:sr_comparison}, SmartDirector achieves PSNR and SSIM scores on par with SparkVSR across all four benchmarks, while delivering substantially lower LPIPS on every dataset. The consistent advantage in perceptual similarity indicates that our method does not merely upscale the low-resolution input, but further restores fine-grained details under the guidance of the high-resolution reference keyframes. 

\begin{table}[htbp]
\centering
\caption{\textbf{Quantitative comparison with SparkVSR on four video super-resolution benchmarks.} Best values per metric are highlighted in \textbf{bold}.}
\label{tab:sr_comparison}
\resizebox{\textwidth}{!}{%
\setlength{\tabcolsep}{6pt}
\renewcommand{\arraystretch}{1.1}
\begin{tabular}{lccccccccccccc}
\toprule
\multirow{2}{*}{\textbf{Method}} & \multicolumn{3}{c}{\textbf{UDM10}} & \multicolumn{3}{c}{\textbf{SPMCS}} & \multicolumn{3}{c}{\textbf{YouHQ40}} & \multicolumn{3}{c}{\textbf{RealVSR}} \\
\cmidrule(lr){2-4} \cmidrule(lr){5-7} \cmidrule(lr){8-10} \cmidrule(lr){11-13}
 & PSNR $\uparrow$ & SSIM $\uparrow$ & LPIPS $\downarrow$ & PSNR $\uparrow$ & SSIM $\uparrow$ & LPIPS $\downarrow$ & PSNR $\uparrow$ & SSIM $\uparrow$ & LPIPS $\downarrow$ & PSNR $\uparrow$ & SSIM $\uparrow$ & LPIPS $\downarrow$ \\
\midrule
SparkVSR & \textbf{23.43} & \textbf{0.6710} & 0.3548 & 20.12 & 0.4908 & 0.3387 & 21.75 & 0.5786 & 0.3501 & \textbf{19.72} & \textbf{0.6183} & 0.2165 \\
SmartDirector & 22.78 & 0.6357 & \textbf{0.2016} & \textbf{21.01} & \textbf{0.5207} & \textbf{0.2235} & \textbf{22.12} & \textbf{0.5824} & \textbf{0.1366} & 18.44 & 0.6033 & \textbf{0.1462} \\
\bottomrule
\end{tabular}
}
\end{table}

\begin{figure}[htbp]
    \centering
    \includegraphics[width=\linewidth]{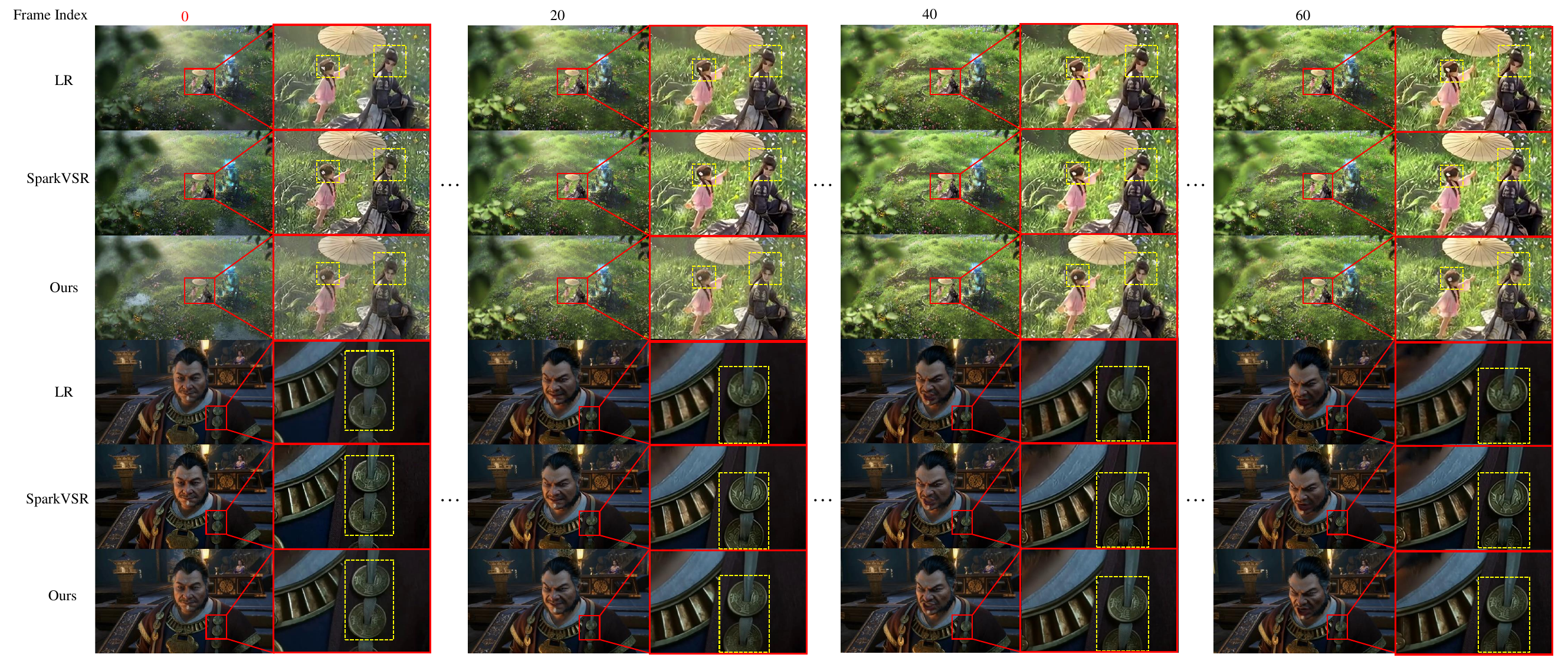}
    \caption{\textbf{Qualitative comparison with SparkVSR.} SmartDirector outperforms SparkVSR on challenging cases such as distorted facial structures and corrupted textual content.}
    \label{fig:sr_compare}
    \vspace{-12pt}
\end{figure}

\textbf{Qualitative Results. }
We further provide qualitative comparisons between SmartDirector and SparkVSR in Fig.~\ref{fig:sr_compare}. Although both methods are conditioned on high-resolution keyframes, SparkVSR struggles to recover severely degraded regions in the low-resolution inputs, such as distorted facial structures and corrupted textual content. In contrast, SmartDirector successfully eliminates these artifacts and reconstructs faithful facial details and legible text that are visually consistent with the high-resolution keyframes. 
This advantage stems from our Multi-Chunk VAE strategy, which encodes the HR keyframes and the LR frames independently, so that the HR keyframe latents are not polluted by the neighboring LR frames during VAE encoding. The clean HR keyframe latents thus retain faithful semantic information that can be reliably propagated to the LR regions, yielding substantially better restoration than SparkVSR.

\subsection{Ablation Study}
\begin{figure}[htbp]
    \centering
    \includegraphics[width=\textwidth]{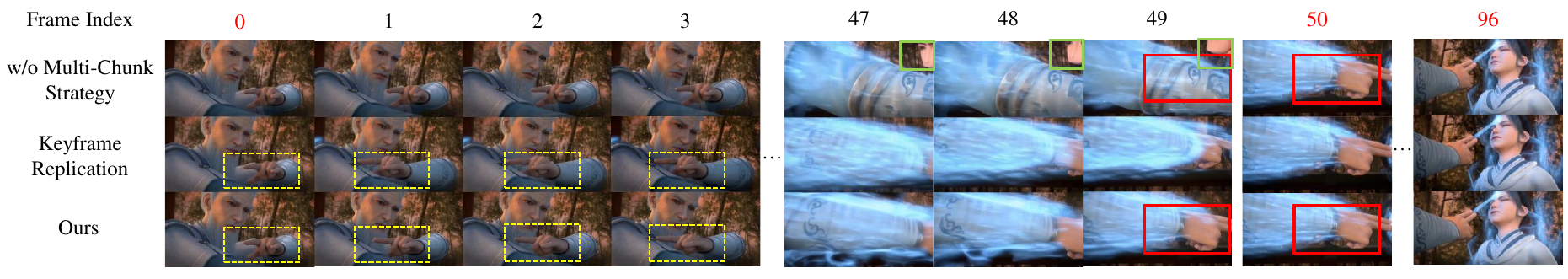}
    \caption{\textbf{Ablation study on keyframe conditioning mechanism.} 
    Compared to two variants, our method maintains both motion fluidity and temporal coherence throughout the sequence. The red frame index denotes the keyframe location.}
    \label{fig:ablation_all}
    \vspace{-12pt}
\end{figure}

To empirically validate the effectiveness of our proposed framework, we perform ablation studies focusing on the \textit{keyframe conditioning mechanism} (Fig.~\ref{fig:ablation_all}). We compare our approach against two variants
\begin{itemize}
    \item w/o Multi-Chunk strategy: this variant directly inserts keyframes into the input latents and violates the VAE causal structure.
    \item Keyframe replication: this variant replicates each keyframe along the temporal axis to fill its corresponding chunk, which introduces temporal redundancy.
\end{itemize}

\textbf{Multi-Chunk strategy.}
As illustrated in Fig.~\ref{fig:ablation_all}, removing the Multi-Chunk strategy leads to severe temporal drift and narrative collapse, which manifests in two distinct failure modes. First, as highlighted by the \emph{red box}, the character's hand exhibits an abrupt motion jump between frames 49 and 50, breaking the temporal trajectory established by the surrounding frames. Second, as highlighted by the \emph{green box}, since this variant violates the causal constraint of the VAE, the model resorts to naive copy-paste from the keyframes: the character at frame 47 is directly replicated from the keyframe at frame 96, and then abruptly disappears at frame 50.
Together, these two observations confirm that directly injecting keyframes into the input latents is fundamentally incompatible with the causal structure of the temporal VAE.

\textbf{Keyframe replication.}
We further examine an alternative that replicates each keyframe four times along the temporal axis and then encodes the replicated frames with the VAE, so that the resulting latents carry only keyframe information while respecting the VAE's causal constraint. However, this design introduces substantial temporal redundancy and leads to noticeable stuttering in the generated video. As highlighted by the \emph{yellow dashed boxes} in Fig.~\ref{fig:ablation_all}, the character's hand remains completely static across consecutive frames, breaking the continuity of motion. In contrast, our multi-chunk design preserves the causal structure while enabling natural temporal dynamics, yielding fluid transitions and coherent narratives.

\vspace{-6pt}

\section{Conclusion}
In this paper, we present SmartDirector, a two-stage framework for keyframe-conditioned cinematic video generation that enables fine-grained control over narrative structure and temporal pacing. The first stage, Director-Gen, synthesizes a low-resolution video from the provided keyframes, and the second stage, Director-SR, refines the result by exploiting high-resolution keyframes as semantic anchors to recover fine-grained details and repair generative artifacts introduced in the first stage. To respect the causal structure of the 3D VAE, we introduce a Multi-Chunk VAE strategy that partitions the latent sequence into keyframe-aligned chunks rather than directly injecting keyframes into the input latents. On top of this representation, we employ full spatio-temporal attention to enable global context exchange across chunks, and further design MC-RoPE to preserve temporal smoothness across chunk boundaries. Extensive experiments on both single-shot and multi-shot scenarios demonstrate that SmartDirector substantially outperforms existing state-of-the-art approaches. Beyond keyframe-conditioned generation, we further show that SmartDirector naturally extends to video extension, highlighting the flexibility and generality of the proposed framework.

{
\small
\bibliographystyle{IEEEtran}
\bibliography{main}

@String(CVPRW = {IEEE Conf. Comput. Vis. Pattern Recog. Worksh.})

@String(CVPRW = {CVPRW})

@inproceedings{xiao2025captain,
  title={Captain cinema: Towards short movie generation},
  author={Xiao, Junfei and Yang, Ceyuan and Zhang, Lvmin and Cai, Shengqu and Zhao, Yang and Guo, Yuwei and Wetzstein, Gordon and Agrawala, Maneesh and Yuille, Alan and Jiang, Lu},
  booktitle={The Fourteenth International Conference on Learning Representations},
  year={2025}
}

@article{liu2025dreamontage,
  title={DreaMontage: Arbitrary Frame-Guided One-Shot Video Generation},
  author={Liu, Jiawei and Li, Junqiao and Deng, Jiangfan and Li, Gen and Zhou, Siyu and Fang, Zetao and Lao, Shanshan and Deng, Zengde and Zhu, Jianing and Ma, Tingting and others},
  journal={arXiv preprint arXiv:2512.21252},
  year={2025}
}

@article{wang2025multishotmaster,
  title={MultiShotMaster: A Controllable Multi-Shot Video Generation Framework},
  author={Wang, Qinghe and Shi, Xiaoyu and Li, Baolu and Bian, Weikang and Liu, Quande and Lu, Huchuan and Wang, Xintao and Wan, Pengfei and Gai, Kun and Jia, Xu},
  journal={arXiv preprint arXiv:2512.03041},
  year={2025}
}

@article{sora,
  title={Video generation models as world simulators},
  author={Tim Brooks and Bill Peebles and Connor Holmes and Will DePue and Yufei Guo and Li Jing and David Schnurr and Joe Taylor and Troy Luhman and Eric Luhman and Clarence Ng and Ricky Wang and Aditya Ramesh},
  journal={OpenAI Technical Report},
  year={2024},
  url={https://openai.com/research/video-generation-models-as-world-simulators},
}

@article{veo,
  title={Veo: Our most capable generative video model},
  author={Google DeepMind.},
  journal={Google DeepMind Blog},
  year={2024},
  url={https://deepmind.google/technologies/veo/},
}

@misc{klingteam2025klingomnitechnicalreport,
      title={Kling-Omni Technical Report}, 
      author={Kling Team and Jialu Chen and Yuanzheng Ci and Xiangyu Du and Zipeng Feng and Kun Gai and Sainan Guo and Feng Han and Jingbin He and Kang He and Xiao Hu and Xiaohua Hu and Boyuan Jiang and Fangyuan Kong and Hang Li and Jie Li and Qingyu Li and Shen Li and Xiaohan Li and Yan Li and Jiajun Liang and Borui Liao and Yiqiao Liao and Weihong Lin and Quande Liu and Xiaokun Liu and Yilun Liu and Yuliang Liu and Shun Lu and Hangyu Mao and Yunyao Mao and Haodong Ouyang and Wenyu Qin and Wanqi Shi and Xiaoyu Shi and Lianghao Su and Haozhi Sun and Peiqin Sun and Pengfei Wan and Chao Wang and Chenyu Wang and Meng Wang and Qiulin Wang and Runqi Wang and Xintao Wang and Xuebo Wang and Zekun Wang and Min Wei and Tiancheng Wen and Guohao Wu and Xiaoshi Wu and Zhenhua Wu and Da Xie and Yingtong Xiong and Yulong Xu and Sile Yang and Zikang Yang and Weicai Ye and Ziyang Yuan and Shenglong Zhang and Shuaiyu Zhang and Yuanxing Zhang and Yufan Zhang and Wenzheng Zhao and Ruiliang Zhou and Yan Zhou and Guosheng Zhu and Yongjie Zhu},
      year={2025},
      eprint={2512.16776},
      archivePrefix={arXiv},
      primaryClass={cs.CV},
      url={https://arxiv.org/abs/2512.16776}, 
}

@misc{wan2025wanopenadvancedlargescale,
      title={Wan: Open and Advanced Large-Scale Video Generative Models}, 
      author={Team Wan and Ang Wang and Baole Ai and Bin Wen and Chaojie Mao and Chen-Wei Xie and Di Chen and Feiwu Yu and Haiming Zhao and Jianxiao Yang and Jianyuan Zeng and Jiayu Wang and Jingfeng Zhang and Jingren Zhou and Jinkai Wang and Jixuan Chen and Kai Zhu and Kang Zhao and Keyu Yan and Lianghua Huang and Mengyang Feng and Ningyi Zhang and Pandeng Li and Pingyu Wu and Ruihang Chu and Ruili Feng and Shiwei Zhang and Siyang Sun and Tao Fang and Tianxing Wang and Tianyi Gui and Tingyu Weng and Tong Shen and Wei Lin and Wei Wang and Wei Wang and Wenmeng Zhou and Wente Wang and Wenting Shen and Wenyuan Yu and Xianzhong Shi and Xiaoming Huang and Xin Xu and Yan Kou and Yangyu Lv and Yifei Li and Yijing Liu and Yiming Wang and Yingya Zhang and Yitong Huang and Yong Li and You Wu and Yu Liu and Yulin Pan and Yun Zheng and Yuntao Hong and Yupeng Shi and Yutong Feng and Zeyinzi Jiang and Zhen Han and Zhi-Fan Wu and Ziyu Liu},
      year={2025},
      eprint={2503.20314},
      archivePrefix={arXiv},
      primaryClass={cs.CV},
      url={https://arxiv.org/abs/2503.20314}, 
}

@article{kong2024hunyuanvideo,
  title={Hunyuanvideo: A systematic framework for large video generative models},
  author={Kong, Weijie and Tian, Qi and Zhang, Zijian and Min, Rox and Dai, Zuozhuo and Zhou, Jin and Xiong, Jiangfeng and Li, Xin and Wu, Bo and Zhang, Jianwei and others},
  journal={arXiv preprint arXiv:2412.03603},
  year={2024}
}

@article{yang2024cogvideox,
  title={CogVideoX: Text-to-Video Diffusion Models with An Expert Transformer},
  author={Yang, Zhuoyi and Teng, Jiayan and Zheng, Wendi and Ding, Ming and Huang, Shiyu and Xu, Jiazheng and Yang, Yuanming and Hong, Wenyi and Zhang, Xiaohan and Feng, Guanyu and others},
  journal={arXiv preprint arXiv:2408.06072},
  year={2024}
}

@article{HaCohen2024LTXVideo,
  title={LTX-Video: Realtime Video Latent Diffusion},
  author={HaCohen, Yoav and Chiprut, Nisan and Brazowski, Benny and Shalem, Daniel and Moshe, Dudu and Richardson, Eitan and Levin, Eran and Shiran, Guy and Zabari, Nir and Gordon, Ori and Panet, Poriya and Weissbuch, Sapir and Kulikov, Victor and Bitterman, Yaki and Melumian, Zeev and Bibi, Ofir},
  journal={arXiv preprint arXiv:2501.00103},
  year={2024}
}

@inproceedings{chan2022investigating,
  title={Investigating tradeoffs in real-world video super-resolution},
  author={Chan, Kelvin CK and Zhou, Shangchen and Xu, Xiangyu and Loy, Chen Change},
  booktitle={Proceedings of the IEEE/CVF conference on computer vision and pattern recognition},
  pages={5962--5971},
  year={2022}
}

@article{chen2025dove,
  title={Dove: Efficient one-step diffusion model for real-world video super-resolution},
  author={Chen, Zheng and Zou, Zichen and Zhang, Kewei and Su, Xiongfei and Yuan, Xin and Guo, Yong and Zhang, Yulun},
  journal={arXiv preprint arXiv:2505.16239},
  year={2025}
}

@article{zhuang2025flashvsr,
  title={Flashvsr: Towards real-time diffusion-based streaming video super-resolution},
  author={Zhuang, Junhao and Guo, Shi and Cai, Xin and Li, Xiaohui and Liu, Yihao and Yuan, Chun and Xue, Tianfan},
  journal={arXiv preprint arXiv:2510.12747},
  year={2025}
}

@inproceedings{wang2025seedvr,
  title={Seedvr: Seeding infinity in diffusion transformer towards generic video restoration},
  author={Wang, Jianyi and Lin, Zhijie and Wei, Meng and Zhao, Yang and Yang, Ceyuan and Loy, Chen Change and Jiang, Lu},
  booktitle={Proceedings of the IEEE/CVF Conference on Computer Vision and Pattern Recognition},
  pages={2161--2172},
  year={2025}
}

@misc{jimeng2024,
  author = {ByteDance},
  title = {Jimeng AI},
  year = {2024},
  howpublished = {\url{https://jimeng.jianying.com/}},
}

@article{team2023gemini,
  title={Gemini: a family of highly capable multimodal models},
  author={Team, Gemini and Anil, Rohan and Borgeaud, Sebastian and Alayrac, Jean-Baptiste and Yu, Jiahui and Soricut, Radu and Schalkwyk, Johan and Dai, Andrew M and Hauth, Anja and Millican, Katie and others},
  journal={arXiv preprint arXiv:2312.11805},
  year={2023}
}

@inproceedings{esser2024scaling,
  title={Scaling rectified flow transformers for high-resolution image synthesis},
  author={Esser, Patrick and Kulal, Sumith and Blattmann, Andreas and Entezari, Rahim and M{\"u}ller, Jonas and Saini, Harry and Levi, Yam and Lorenz, Dominik and Sauer, Axel and Boesel, Frederic and others},
  booktitle={Forty-first international conference on machine learning},
  year={2024}
}

@misc{flux2024,
    author={Black Forest Labs},
    title={FLUX},
    year={2024},
    howpublished={\url{https://github.com/black-forest-labs/flux}},
}

@inproceedings{chen2025goku,
  title={Goku: Flow based video generative foundation models},
  author={Chen, Shoufa and Ge, Chongjian and Zhang, Yuqi and Zhang, Yida and Zhu, Fengda and Yang, Hao and Hao, Hongxiang and Wu, Hui and Lai, Zhichao and Hu, Yifei and others},
  booktitle={Proceedings of the Computer Vision and Pattern Recognition Conference},
  pages={23516--23527},
  year={2025}
}

@article{liu2022flow,
  title={Flow straight and fast: Learning to generate and transfer data with rectified flow},
  author={Liu, Xingchao and Gong, Chengyue and Liu, Qiang},
  journal={arXiv preprint arXiv:2209.03003},
  year={2022}
}

@article{yu2026sparkvsr,
  title={SparkVSR: Interactive Video Super-Resolution via Sparse Keyframe Propagation},
  author={Yu, Jiongze and Gao, Xiangbo and Verlani, Pooja and Gadde, Akshay and Wang, Yilin and Adsumilli, Balu and Tu, Zhengzhong},
  journal={arXiv preprint arXiv:2603.16864},
  year={2026}
}

@article{meng2025holocine,
  title={Holocine: Holistic generation of cinematic multi-shot long video narratives},
  author={Meng, Yihao and Ouyang, Hao and Yu, Yue and Wang, Qiuyu and Wang, Wen and Cheng, Ka Leong and Wang, Hanlin and Li, Yixuan and Chen, Cheng and Zeng, Yanhong and others},
  journal={arXiv preprint arXiv:2510.20822},
  year={2025}
}

@article{liu2025pusa,
  title={Pusa v1. 0: Surpassing wan-i2v with \$500 training cost by vectorized timestep adaptation},
  author={Liu, Yaofang and Ren, Yumeng and Artola, Aitor and Hu, Yuxuan and Cun, Xiaodong and Zhao, Xiaotong and Zhao, Alan and Chan, Raymond H and Zhang, Suiyun and Liu, Rui and others},
  journal={arXiv preprint arXiv:2507.16116},
  year={2025}
}

@inproceedings{wang2025vggt,
  title={Vggt: Visual geometry grounded transformer},
  author={Wang, Jianyuan and Chen, Minghao and Karaev, Nikita and Vedaldi, Andrea and Rupprecht, Christian and Novotny, David},
  booktitle={Proceedings of the Computer Vision and Pattern Recognition Conference},
  pages={5294--5306},
  year={2025}
}

@inproceedings{zhuautoshot,
  title={AutoShot: A Short Video Dataset and State-of-the-Art Shot Boundary Detection},
  author={Zhu, Wentao and Huang, Yufang and Xie, Xiufeng and Liu, Wenxian and Deng, Jincan and Zhang, Debing and Wang, Zhangyang and Liu, Ji},
  booktitle={Proceedings of the IEEE/CVF Conference on Computer Vision and Pattern Recognition Workshops (CVPRW)},
  year={2023}
}

@article{ravi2024sam2,
  title={SAM 2: Segment Anything in Images and Videos},
  author={Ravi, Nikhila and Gabeur, Valentin and Hu, Yuan-Ting and Hu, Ronghang and Ryali, Chaitanya and Ma, Tengyu and Khedr, Haitham and R{\"a}dle, Roman and Rolland, Chloe and Gustafson, Laura and Mintun, Eric and Pan, Junting and Alwala, Kalyan Vasudev and Carion, Nicolas and Wu, Chao-Yuan and Girshick, Ross and Doll{\'a}r, Piotr and Feichtenhofer, Christoph},
  journal={arXiv preprint arXiv:2408.00714},
  url={https://arxiv.org/abs/2408.00714},
  year={2024}
}

@article{Qwen3-VL,
      title={Qwen3-VL Technical Report}, 
      author={Shuai Bai and Yuxuan Cai and Ruizhe Chen and Keqin Chen and Xionghui Chen and Zesen Cheng and Lianghao Deng and Wei Ding and Chang Gao and Chunjiang Ge and Wenbin Ge and Zhifang Guo and Qidong Huang and Jie Huang and Fei Huang and Binyuan Hui and Shutong Jiang and Zhaohai Li and Mingsheng Li and Mei Li and Kaixin Li and Zicheng Lin and Junyang Lin and Xuejing Liu and Jiawei Liu and Chenglong Liu and Yang Liu and Dayiheng Liu and Shixuan Liu and Dunjie Lu and Ruilin Luo and Chenxu Lv and Rui Men and Lingchen Meng and Xuancheng Ren and Xingzhang Ren and Sibo Song and Yuchong Sun and Jun Tang and Jianhong Tu and Jianqiang Wan and Peng Wang and Pengfei Wang and Qiuyue Wang and Yuxuan Wang and Tianbao Xie and Yiheng Xu and Haiyang Xu and Jin Xu and Zhibo Yang and Mingkun Yang and Jianxin Yang and An Yang and Bowen Yu and Fei Zhang and Hang Zhang and Xi Zhang and Bo Zheng and Humen Zhong and Jingren Zhou and Fan Zhou and Jing Zhou and Yuanzhi Zhu and Ke Zhu},
	  journal={arXiv preprint arXiv:2511.21631},
      year={2025}
}

@article{lipman2022flow,
  title={Flow matching for generative modeling},
  author={Lipman, Yaron and Chen, Ricky TQ and Ben-Hamu, Heli and Nickel, Maximilian and Le, Matt},
  journal={arXiv preprint arXiv:2210.02747},
  year={2022}
}

@misc{wiki:storyboard,
  author       = {{Wikipedia contributors}},
  title        = {Storyboard --- {W}ikipedia{,} The Free Encyclopedia},
  howpublished = {\url{https://en.wikipedia.org/wiki/Storyboard}},
  year         = {2026},
  note         = {[Online; accessed 6-May-2026]}
}

@article{xue2025ultravideo,
  title={Ultravideo: High-quality uhd video dataset with comprehensive captions},
  author={Xue, Zhucun and Zhang, Jiangning and Hu, Teng and He, Haoyang and Chen, Yinan and Cai, Yuxuan and Wang, Yabiao and Wang, Chengjie and Liu, Yong and Li, Xiangtai and others},
  journal={arXiv preprint arXiv:2506.13691},
  year={2025}
}

@inproceedings{peebles2023scalable,
  title={Scalable diffusion models with transformers},
  author={Peebles, William and Xie, Saining},
  booktitle={Proceedings of the IEEE/CVF international conference on computer vision},
  pages={4195--4205},
  year={2023}
}

@article{chu2018temporally,
  title={Temporally coherent gans for video super-resolution (tecogan)},
  author={Chu, Mengyu and Xie, You and Leal-Taix{\'e}, Laura and Thuerey, Nils},
  journal={arXiv preprint arXiv:1811.09393},
  volume={1},
  number={2},
  pages={3},
  year={2018}
}
}

\newpage

\section{Appendix: LLM Evaluation Protocol}
\label{sec:appendix_llm_eval}

We provide the complete system prompt used for the Gemini-based evaluation below. The evaluator is instructed to perform blind visual analysis followed by prompt-consistency checking, outputting results in a structured JSON format.

\begin{tcolorbox}[
    colback=gray!5!white, 
    colframe=black, 
    sharp corners, 
    boxrule=0.8pt,
    title=\textbf{Prompt For Instruction-Following Assessment}
]
\small
\vspace{-0.5em}
\textbf{Task Definition}

You will be provided with a text prompt (describing the target sequence, characters, and environment), a set of reference keyframes extracted from a source video, and a generated video conditioned on these inputs. 
Your task is to evaluate how accurately the generated video adheres to the text prompt while preserving the semantic and spatial information anchored by the reference keyframes.
Specifically, you must assess the video's consistency and fidelity across four sub-dimensions: Character Consistency, Emotion Consistency, Scene Consistency, and Spatial Consistency.

\textbf{Evaluation Criteria}
\vspace{-0.5em}

\begin{itemize}[leftmargin=*]
    \item \textbf{Character Consistency}
    \begin{itemize}
        \item \textbf{Definition}: Evaluate whether the characters' identity, physical features (hair, eyes, skin tone), and attire remain stable across all keyframes and align with the text description.
        \item \textbf{Key Points}: Are there sudden changes in clothing or facial features? Does the character look like the same person from start to finish?
    \end{itemize}
    \item \textbf{Emotion Consistency}
    \begin{itemize}
        \item \textbf{Definition}: Assess if the character's facial expressions and body language accurately reflect the mood or emotional arc described in the prompt.
        \item \textbf{Key Points}: Does the emotion match the text? If the prompt describes a transition (e.g., from sad to happy), is that transition captured logically across the frames?
    \end{itemize}
    \item \textbf{Scene Consistency}
    \begin{itemize}
        \item \textbf{Definition}: Evaluate whether the background, environmental elements, lighting, and props match the text description and remain stable across the sequence.
        \item \textbf{Key Points}: Does the setting change abruptly without reason? Are the objects mentioned in the prompt present and visually consistent?
    \end{itemize}
    \item \textbf{Spatial Consistency}
    \begin{itemize}
        \item \textbf{Definition}: Assess the logical positioning of characters and objects within the 3D space, as well as the camera perspective described in the prompt.
        \item \textbf{Key Points}: Does the relative distance between objects make sense? Does the movement or camera angle follow the prompt's spatial instructions (e.g., "panning left," "close-up")?
    \end{itemize}
\end{itemize}

\vspace{-0.2em}
\textbf{Scoring Criteria}

Assign a score from 0 to 5 based on the overall performance across all four dimensions:
\begin{itemize}[leftmargin=*]
    \item \textbf{Unacceptable (0)}: Completely fails expectations.
    \item \textbf{Very Poor (1)}: No Correlation. The video sequence fails to reflect the prompt in almost all dimensions. Complete lack of consistency.
    \item \textbf{Poor (2)}: Weak Correlation. Only minor elements match the prompt. Significant "hallucinations" or jarring inconsistencies in characters or environment.
    \item \textbf{Fair (3)}: Moderate Correlation. The main idea is present, but there are noticeable flaws (e.g., character's clothes change color, background shifts unnaturally, or emotion is mismatched).
    \item \textbf{Good (4)}: Strong Correlation. High fidelity to the prompt. Most dimensions are well-maintained with only very minor flickering or slight spatial deviations.
    \item \textbf{Excellent (5)}: Near-Perfect Correlation. Perfect instruction following. Characters, emotions, scenes, and spatial logic are flawlessly executed as described.
\end{itemize}

\textbf{Input Format}
\begin{itemize}[leftmargin=*]
    \vspace{-0.2em}
    \item \textbf{Generated Video}: [The video file produced by the generation model]
    \vspace{-0.2em}
    \item \textbf{Text Prompt}: [Detailed description of the video content]
    \vspace{-0.2em}
    \item \textbf{Keyframes}: [A series of images representing the video sequence]
\end{itemize}

\textbf{Output Format}

Please provide a detailed analysis for each dimension, followed by an overall summary and a final score.

\textbf{Score}: <0-5>
\end{tcolorbox}

\begin{tcolorbox}[
    colback=gray!5!white, 
    colframe=black, 
    sharp corners, 
    boxrule=0.8pt,
    title=\textbf{Prompt For Video Narrative Coherence Assessment}
]
\small
\textbf{Task Definition}

You will be provided with a text prompt (describing the target sequence, characters, and environment), a set of reference keyframes extracted from a source video, and a generated video conditioned on these inputs. 
Your task is to evaluate the Narrative Coherence of the video. This involves assessing how logically the story unfolds across the frames and whether the movements/actions within the sequence adhere to physical and causal laws.

\textbf{Evaluation Criteria}

When assessing the narrative Coherence, focus on the following two dimensions:
\begin{itemize}[leftmargin=*]
    \item \textbf{Plot Coherence}
    \begin{itemize}
        \item \textbf{Flow of Events}: Evaluate whether the sequence of frames tells a clear, continuous story that aligns with the prompt. There should be a logical progression from the beginning to the end.
        \item \textbf{Narrative Completeness}: Does the video capture the key story beats described in the prompt? No critical narrative steps should be skipped (e.g., if the prompt says "a man sits down," we should see the transition, not just him standing then suddenly sitting).
        \item \textbf{Causal Logic}: Do events follow a cause-and-effect relationship? (e.g., if a character throws a ball, the next frames should show the ball in flight).
    \end{itemize}
    \item \textbf{Action Logic}
    \begin{itemize}
        \item \textbf{Physical Realism}: Evaluate whether the movements of characters and objects follow the laws of physics. Actions should not look "glitched," teleported, or physically impossible unless specified by the prompt.
        \item \textbf{Motion Continuity}: The direction and momentum of movement must be consistent across frames. (e.g., if a character is running left, they should not suddenly be moving right in the next frame without a visible turn).
        \item \textbf{Interaction Logic}: When characters interact with objects or the environment, the interaction should be plausible (e.g., hands properly gripping tools, feet touching the ground correctly).
    \end{itemize}
\end{itemize}

\textbf{Scoring Criteria}

Based on the narrative flow and action logic, assign a score from 0 to 5:
\begin{itemize}[leftmargin=*]
    \item \textbf{Unacceptable (0)}: Completely fails expectations.
    \item \textbf{Very Poor (1)}: No Narrative Logic. The frames are a disconnected collection of images with no storytelling or logical movement. Actions are physically impossible or chaotic.
    \item \textbf{Poor (2)}: Weak Narrative. The sequence captures the general idea but has major "teleportation" issues, missing narrative steps, or jarringly illogical movements.
    \item \textbf{Fair (3)}: Moderate Narrative. The basic story is recognizable, but there are noticeable "jumps" in the plot or minor physical glitches in how actions are performed.
    \item \textbf{Good (4)}: Strong Narrative. The story flows well and aligns closely with the prompt. Actions are mostly logical and smooth, with only very minor pacing issues or slight physical inconsistencies.
    \item \textbf{Excellent (5)}: Near-Perfect Narrative. The sequence perfectly executes the described story with flawless action logic, smooth transitions, and complete causal consistency.
\end{itemize}

\textbf{Input Format}
\begin{itemize}[leftmargin=*]
    \item \textbf{Generated Video}: [The video file produced by the generation model]
    \item \textbf{Text Prompt}: [Detailed description of the video content]
    \item \textbf{Keyframes}: [A series of images representing the video sequence]
\end{itemize}

\textbf{Output Format}

Please provide a detailed analysis for each dimension, followed by an overall summary and a final score.

\textbf{Score}: <0-5>
\end{tcolorbox}

\begin{tcolorbox}[
    colback=gray!5!white, 
    colframe=black, 
    sharp corners, 
    boxrule=0.8pt,
    title=\textbf{Prompt For Video Physical Consistency Assessment}
]
\small
\textbf{Task Definition}

You will be provided with a text prompt (describing the target sequence, characters, and environment), a set of reference keyframes extracted from a source video, and a generated video conditioned on these inputs. 
Your task is to evaluate the Physical Consistency of the visual content. You must determine if the movements, interactions, and environmental behaviors in the video adhere to the fundamental laws of physics and common sense of the real world.

\textbf{Evaluation Criteria}

When assessing Physical Consistency, evaluate the sequence based on the following aspects:
\begin{itemize}[leftmargin=*]
    \item \textbf{Gravity and Weight}: Do objects and characters fall, jump, or rest with a realistic sense of gravity? There should be no unnatural floating or "weightless" movements unless specified by the prompt.
    \item \textbf{Collision and Clipping}: Check if solid objects pass through each other inappropriately (clipping). Interactions between objects (e.g., a hand grabbing a cup, feet touching the ground) should show proper contact and resistance without merging textures.
    \item \textbf{Object Permanence and Structural Integrity}: Objects should not spontaneously disappear, morph into other shapes, or dissolve into the background. The structural integrity of complex objects (like bicycles, limbs, or tools) must remain logical during movement.
    \item \textbf{Fluid and Particle Dynamics}: If present, evaluate the behavior of water, fire, smoke, or hair. Their movement should follow realistic patterns (e.g., water flowing downward, hair reacting to wind or movement).
    \item \textbf{Inertia and Momentum}: Movements should have realistic acceleration and deceleration. Sudden stops or instant changes in direction without corresponding force should be penalized.
\end{itemize}

\textbf{Scoring Criteria}

Based on the adherence to physical laws, assign a score from 0 to 5:
\begin{itemize}[leftmargin=*]
    \item \textbf{Unacceptable (0)}: Completely fails expectations.
    \item \textbf{Very Poor (1)}: Complete Physical Collapse. The video exhibits extreme physical errors, such as objects teleporting, characters melting, or massive clipping that makes the scene incomprehensible.
    \item \textbf{Poor (2)}: Significant Physical Violations. Major physical glitches are present (e.g., a person walking through a wall, objects floating randomly, or limbs bending in impossible ways).
    \item \textbf{Fair (3)}: Noticeable Physical Errors. The basic physics are present, but there are clear "AI hallucinations" like minor clipping, hair/clothing moving independently of physics, or a lack of realistic weight.
    \item \textbf{Good (4)}: Minor Physical Inconsistencies. The physics are mostly believable. Only very subtle errors exist, such as slightly unnatural fluid movement or minor contact issues that do not break the overall realism.
    \item \textbf{Excellent (5)}: Near-Perfect Physical Realism. The video perfectly simulates real-world physics. Interactions, gravity, and object permanence are flawless and look like a real-world recording.
\end{itemize}

\textbf{Input Format}
\begin{itemize}[leftmargin=*]
    \item \textbf{Generated Video}: [The video file produced by the generation model]
    \item \textbf{Text Prompt}: [Detailed description of the video content]
    \item \textbf{Keyframes}: [A series of images representing the video sequence]
\end{itemize}

\textbf{Output Format}

Please provide a detailed analysis for each dimension, followed by an overall summary and a final score.

\textbf{Score}: <0-5>
\end{tcolorbox}

\begin{tcolorbox}[
    colback=gray!5!white, 
    colframe=black, 
    sharp corners, 
    boxrule=0.8pt,
    title=\textbf{Prompt For Video Quality Assessment (Visual Artifacts)}
]
\small
\textbf{Task Definition}

You will be provided with a text prompt (describing the target sequence, characters, and environment), a set of reference keyframes extracted from a source video, and a generated video conditioned on these inputs. 
Your task is to evaluate the visual quality of the generated content, focusing specifically on the presence and severity of visual artifacts. You must assess the clarity, stability, and overall technical execution of the video generation process.

\textbf{Evaluation Criteria}

When assessing visual quality, evaluate the sequence based on the following aspects:
\begin{itemize}[leftmargin=*]
    \item \textbf{Flickering and Strobe}: Check for unstable brightness, textures, or geometry that change rapidly between frames, causing visual discomfort.
    \item \textbf{Blurriness and Noise}: Evaluate the sharpness of the image. Are there excessive compression artifacts, grain, or motion blur that obscures important details?
    \item \textbf{Geometric Distortion}: Check for warped shapes or unnatural stretching of objects and characters, particularly during movement.
    \item \textbf{Color Consistency}: Evaluate whether the color palette remains stable or if there are sudden shifts in lighting/tone that break immersion.
    \item \textbf{Resolution and Detail}: Are fine details (e.g., facial features, textures, small objects) preserved, or do they dissolve into "AI mush" or blobs?
\end{itemize}

\textbf{Scoring Criteria}

Based on the visual quality and frequency of artifacts, assign a score from 0 to 5:
\begin{itemize}[leftmargin=*]
    \item \textbf{Unacceptable (0)}: Completely fails expectations.
    \item \textbf{Very Poor (1)}: Extreme artifacts. Massive flickering, constant blur, or severe geometric distortion throughout the sequence.
    \item \textbf{Poor (2)}: Significant artifacts. Frequent flickering, noticeable compression noise, or obvious warping that distracts from the content.
    \item \textbf{Fair (3)}: Moderate artifacts. Visible issues like occasional flickering or minor loss of detail, but the main content remains viewable.
    \item \textbf{Good (4)}: Minor artifacts. High visual quality with only very subtle imperfections or slight flickering in complex areas.
    \item \textbf{Excellent (5)}: Pristine visual quality. No noticeable artifacts, sharp textures, stable colors, and high-fidelity rendering.
\end{itemize}

\textbf{Input Format}
\begin{itemize}[leftmargin=*]
    \item \textbf{Generated Video}: [The video file produced by the generation model]
    \item \textbf{Text Prompt}: [Detailed description of the video content]
    \item \textbf{Keyframes}: [A series of images representing the video sequence]
\end{itemize}

\textbf{Output Format}

Please provide a detailed analysis for each dimension, followed by an overall summary and a final score.

\textbf{Score}: <0-5>
\end{tcolorbox}

\begin{tcolorbox}[
    colback=gray!5!white, 
    colframe=black, 
    sharp corners, 
    boxrule=0.8pt,
    title=\textbf{Prompt For Video Aesthetic Quality Assessment}
]
\small
\textbf{Task Definition}

You will be provided with a text prompt (describing the target sequence, characters, and environment), a set of reference keyframes extracted from a source video, and a generated video conditioned on these inputs. 
Your task is to evaluate the Aesthetic Quality of the video. You will assess how visually appealing the sequence is and how well the artistic elements—specifically lighting, color, and composition—align with the intended atmosphere described in the prompt.

\textbf{Evaluation Criteria}

When assessing the aesthetic quality, focus on the following three dimensions:
\begin{itemize}[leftmargin=*]
    \item \textbf{Lighting and Shadow}
    \begin{itemize}
        \item \textbf{Directionality \& Depth}: Does the lighting have a clear direction? Does it create a sense of three-dimensionality and depth, or is it flat and dull?
        \item \textbf{Atmosphere}: Does the lighting match the mood (e.g., warm golden hour, harsh cinematic shadows, or soft ethereal glow)?
        \item \textbf{Consistency}: Is the lighting logic maintained across all keyframes?
    \end{itemize}
    \item \textbf{Color Palette and Harmony}
    \begin{itemize}
        \item \textbf{Color Scheme}: Evaluate the choice of colors. Are they harmonious and professionally graded?
        \item \textbf{Vibrancy \& Mood}: Do the colors accurately reflect the emotional tone of the prompt (e.g., desaturated for a noir feel, vibrant and saturated for a fantasy setting)?
        \item \textbf{Consistency}: Does the color grading remain stable throughout the sequence?
    \end{itemize}
    \item \textbf{Composition and Framing}
    \begin{itemize}
        \item \textbf{Balance}: Are the subjects and background elements arranged in a visually pleasing way? Consider the use of the rule of thirds, leading lines, or symmetry.
        \item \textbf{Cinematography}: Evaluate the camera angle and framing. Does the composition feel intentional and cinematic, or accidental and cluttered?
        \item \textbf{Visual Focus}: Is the main subject clearly emphasized through framing and depth of field (bokeh effect)?
    \end{itemize}
\end{itemize}

\textbf{Scoring Criteria}

Based on the visual appeal and artistic execution, assign a score from 0 to 5:
\begin{itemize}[leftmargin=*]
    \item \textbf{Unacceptable (0)}: Completely fails expectations.
    \item \textbf{Very Poor (1)}: Visually Chaotic. The video looks amateurish with flat or non-existent lighting, clashing colors, and messy composition. No artistic intent is visible.
    \item \textbf{Poor (2)}: Low Aesthetic Value. The image is bland or unappealing. Lighting is poorly handled, colors are muddy, and the framing feels accidental or awkward.
    \item \textbf{Fair (3)}: Average Quality. The aesthetics are functional but lack professional polish. Lighting and colors are acceptable but generic, and composition is basic.
    \item \textbf{Good (4)}: Visually Pleasing. The sequence has strong cinematic qualities. Lighting creates good depth, the color palette is well-coordinated, and the composition is intentional and effective.
    \item \textbf{Excellent (5)}: Cinematic Masterpiece. The video exhibits professional-grade artistry. Light, color, and composition work together perfectly to create a stunning, immersive visual experience.
\end{itemize}

\textbf{Input Format}
\begin{itemize}[leftmargin=*]
    \item \textbf{Generated Video}: [The video file produced by the generation model]
    \item \textbf{Text Prompt}: [Detailed description of the video content]
    \item \textbf{Keyframes}: [A series of images representing the video sequence]
\end{itemize}

\textbf{Output Format}

Please provide a detailed analysis for each dimension, followed by an overall summary and a final score.

\textbf{Score}: <0-5>
\end{tcolorbox}

\newpage
\section{Detailed Human Evaluation Protocol}
\label{sec:appendix_human_eval}

\subsection{Evaluation Setup}
To ensure a fair and objective assessment, we conducted a blind human evaluation. We randomly sampled videos from our test set and extracted keyframes, which, along with prompts generated by a large language model (LLM), served as inputs for both our \textit{SmartDirector} and the \textit{Dreamina} baseline (multi-frame model).

\subsection{Data Collection}
We deployed a web-based evaluation interface. Each participant was assigned 10 video pairs (5 \textit{Single-Shot} and 5 \textit{Multi-Shot} scenarios). To ensure a double-blind setup, the assignment of "A" and "B" labels was randomized at the backend, ensuring that neither the participants nor the evaluators had prior knowledge of model identities. 

For each pair, participants were required to provide an "Overall Quality" rating. Additionally, they could optionally assess the following three dimensions:
\begin{itemize}
    \item \textbf{Identity Consistency:} Consistency of character IDs, attire, accessories, scene settings, and objects across the temporal timeline and scene transitions.
    \item \textbf{Narrative Pacing:} Coherence and natural rhythm of the narrative progression, assessing whether the timing of events, scene transitions, and temporal flow between keyframes are smooth and logically consistent.
    \item \textbf{Keyframe Adherence:} Accuracy in reconstructing content, composition, and poses from the reference keyframe images.
\end{itemize}

\subsection{Scoring and Statistical Analysis}
The evaluation used a 5-point Likert-style scale: (1) A significantly better than B, (2) A slightly better than B, (3) Similar quality, (4) B slightly better than A, and (5) B significantly better than A. 
We collapsed these into three categories (SmartDirector Better, Neutral, Dreamina Better) to calculate the \textbf{GSB score}, defined as:
\begin{equation}
    GSB = \frac{Wins - Losses}{Wins + Losses + Ties}
\end{equation}
The detailed breakdown of user preferences is presented in Table~\ref{tab:detailed_human_eval}.

\begin{table}[htbp]
\centering
\caption{Detailed human evaluation breakdown (percentage \%) for all dimensions. 'S' and 'D' denote SmartDirector and Dreamina, respectively.}
\label{tab:detailed_human_eval}
\resizebox{\textwidth}{!}{
\begin{tabular}{llccccc}
\toprule
\textbf{Scenario} & \textbf{Dimension} & \textbf{S significantly better} & \textbf{S slightly better} & \textbf{Neutral} & \textbf{D slightly better} & \textbf{D significantly better} \\
\midrule
\multirow{4}{*}{\textbf{All}} 
 & Overall Quality & 39.21 & 23.68 & 16.32 & 8.16 & 12.63 \\
 & Identity Consistency & 25.39 & 25.08 & 38.87 & 4.39 & 6.27 \\
 & Narrative Pacing & 29.02 & 21.14 & 36.59 & 7.57 & 5.68 \\
 & Keyframe Adherence & 26.27 & 27.22 & 34.81 & 5.38 & 6.33 \\
\midrule
\multirow{4}{*}{\textbf{Single}} 
 & Overall Quality & 28.42 & 25.79 & 21.05 & 11.58 & 13.16 \\
 & Identity Consistency & 14.38 & 25.62 & 48.75 & 5.00 & 6.25 \\
 & Narrative Pacing & 14.47 & 20.13 & 50.94 & 10.06 & 4.40 \\
 & Keyframe Adherence & 17.72 & 27.85 & 41.14 & 6.33 & 6.96 \\
\midrule
\multirow{4}{*}{\textbf{Multi}} 
 & Overall Quality & 50.00 & 21.58 & 11.58 & 4.74 & 12.11 \\
 & Identity Consistency & 36.48 & 24.53 & 28.93 & 3.77 & 6.29 \\
 & Narrative Pacing & 43.67 & 22.15 & 22.15 & 5.06 & 6.96 \\
 & Keyframe Adherence & 34.81 & 26.58 & 28.48 & 4.43 & 5.70 \\
\bottomrule
\end{tabular}
}
\end{table}

\section{Limitations}
\label{sec:limitations}
While SmartDirector demonstrates strong capabilities in keyframe-conditioned cinematic video generation, several limitations remain. 

First, while the two-stage pipeline significantly reduces computational overhead compared to direct native 1080p generation, it introduces a recognized trade-off: the initial low-resolution synthesis may slightly weaken fine-grained adherence to keyframe conditioning signals and yield marginally reduced spatial clarity relative to an idealized single-stage high-resolution generator. We explicitly mitigate this gap through our keyframe-conditioned super-resolution stage, which restores high-frequency details and reinforces spatial alignment, though a marginal fidelity gap persists due to the inherent information bottleneck of the initial pass.

Second, the causal structure of the temporal VAE imposes a temporal discretization constraint, requiring each video chunk to align with a $4n+1$ frame structure. This prevents strictly frame-absolute arbitrary keyframe placement. However, the resulting temporal misalignment is strictly bounded within $\pm 2$ frame, which is perceptually negligible and effectively approximates arbitrary temporal conditioning in practical applications.

\end{document}